\documentclass{article} 

\usepackage[final,nonatbib]{nips_2016}
\usepackage[utf8]{inputenc} 
\usepackage[T1]{fontenc}    
\usepackage{url}            
\usepackage{booktabs}       
\usepackage{amsfonts}       
\usepackage{nicefrac}       
\usepackage{microtype}      

\usepackage{multirow}
\usepackage[multidot]{grffile}
\usepackage{times}
\usepackage{latexsym}
\usepackage{amsmath,amsfonts,amssymb}
\usepackage{graphicx}
\usepackage{subfigure}
\usepackage{colortbl}

\newcommand{\comment}[1]{}

\newcommand{\mpmi}{\mbox{pmi}}
\newcommand{\KL}{\mbox{KL}}

\title{PMI Matrix Approximations with Applications to Neural Language Modeling}
\author{
  Oren Melamud \\
  Bar-Ilan University \\
  \texttt{melamuo@cs.biu.ac.il} \\
     \And
   Ido Dagan  \\
   Bar-Ilan University \\
      \texttt{dagan@cs.biu.ac.il} \\
       \And
   Jacob Goldberger  \\
   Bar-Ilan University \\
      \texttt{goldbej@eng.biu.ac.il} \\
  }

\date{}
\begin{document}
\maketitle

\begin{abstract}
The negative sampling (NEG) objective function, used in \emph{word2vec}, is a simplification of the Noise Contrastive Estimation (NCE) method. NEG was found to be highly effective in learning continuous word representations. However, unlike NCE, it was considered inapplicable for the purpose of learning the parameters of a language model.
In this study, we refute this assertion by providing a principled derivation for NEG-based language modeling, founded on a novel analysis of a low-dimensional approximation of the matrix of pointwise mutual information between the contexts and the predicted words.
  The obtained  language modeling is closely related to NCE language models but is based on a simplified objective function.
    We thus provide a unified formulation for two main language processing tasks, namely word embedding and language modeling, based on the NEG objective function.
  Experimental results on two popular language modeling benchmarks show comparable perplexity results, with a small advantage to NEG over NCE.
\end{abstract}

\section{Introduction}

Statistical language models (LMs) are crucial components in many
speech and text processing systems designed for tasks, such
as speech recognition and machine
translation. Language models learn to predict the probability of a word
given a context of preceding words.
Traditional LMs were based on word $n$-grams counts and therefore limited in the scope of the considered context for reasons of data sparsity. However, Recurrent Neural Network (RNN) language models, which can consider arbitrarily long contexts, have shown consistent performance improvements, recently outperforming $n$-gram LMs across a range of tasks.

An important
practical issue associated with neural-network LMs is the high computational
cost incurred. The key factor that limits the scalability of  traditional neural LMs is the computation
of the normalization term in the softmax output layer, whose cost is linearly proportional to the size of the word vocabulary. This has a
significant impact on both training and testing, even when employing modern GPUs, and especially when a
large  vocabulary is used.
Several approaches have been proposed to cope with this scaling issue, including
importance sampling \cite{Bengio2003}, hierarchical softmax \cite{Mnihnips} and
Noise Contrastive Estimation (NCE) \cite{Gutmann2012}.
NCE reduces the language model estimation problem to the problem of estimating the parameters of a probabilistic
binary classifier that distinguishes between samples from the empirical distribution and samples generated by the noise distribution. This method significantly reduces training time, by making the cost of training independent of the size of the vocabulary, yet still requires the expensive normalization at test time.
NCE has been applied to train neural LMs with large vocabularies \cite{Mnih2012} and was also recently successfully used to train LSTM-RNN LMs
(see e.g.  \cite{Vaswani2013} \cite{Chen2015} \cite{Zoph2016}).

In a related research line, continuous word embeddings have proven useful in many NLP tasks.
In particular, the skip-gram embedding model with the negative sampling (NEG) objective function \cite{Mikolov_nips} as implemented in the \emph{word2vec} toolkit, has become one of the most popular models today. This is mainly attributed to its scalability to large volumes of data, which is critical for learning high-quality embeddings. Indeed, recent studies have obtained state-of-the-art results by using skip-gram embeddings on a variety of natural language processing tasks, such as named entity resolution and dependency parsing \cite{Passos,bansal2014,melamud2016therole}.
The same embedding approach can be used for sentence representation  \cite{Kiros} and context representation \cite{melamud2016conll}.
Recently, Levy and Goldberg \cite{Levy_nips} offered some motivation for skip-gram's NEG objective function, showing that by maximizing this function the skip-gram algorithm implicitly factorizes a word-context pointwise mutual information (PMI) matrix.

Interestingly, the NEG objective function, commonly used for learning word embeddings, is a simplified version of the respective NCE function that is used in language modeling.

The main difference is
that while both NCE and NEG objective functions include the neural representations of context and predicted words, NCE includes in addition the numerical word probabilities in the
noise distribution. Dyer \cite{Dyer2014} argued  that although NEG and NCE are superficially similar, NCE is a general parameter estimation technique that is asymptotically
unbiased, while NEG is most useful for learning word representations, but not as a general-purpose estimator.  In other words, the claim was that the simplified NEG objective is not applicable for language modeling tasks, where you want the output layer to be a probability distribution.

In this study we show that the NEG objective function, in spite of its simplicity, is suitable for training an unbiased language model estimator.
We present a derivation of a NEG-based language modeling algorithm that is founded on an extension of the observation of Levy and Goldberg \cite{Levy_nips}, which showed the relation between the skip-gram algorithm and PMI matrix factorization. We thus  provide a  unified formulation for two main language processing tasks, namely word embedding and language modeling, based on the NEG objective function.

The obtained NEG language modeling algorithm
can be viewed as a variant of the NCE algorithm. It has a  simplified objective function formulation which allows it to avoid heuristic components and initialization procedures that are used in various implementation of NCE language models \cite{Vaswani2013} \cite{Chen2015} \cite{Zoph2016}. We compare NEG to NCE language modeling by evaluating their perplexity measures on two standard language model datasets with different training data and vocabulary size. Our results show that they perform comparably with a small advantage to NEG over NCE.

The remainder of this paper is organized as follows. In Section \ref{sec:2} we provide a formal derivation for how learning Euclidean embeddings with the NEG objective function can be used to approximate any discrete joint distribution by means of PMI. In Section \ref{sec:3} we apply this general principle to language modeling and formalize our proposed NEG LMs. In Section \ref{sec:4} we describe the relation of NEG LMs to NCE LMs, and in Section \ref{sec:5} we compare empirically the performance of LSTM-based NEG and NCE language model implementations.

\section{Euclidean Embedding of a Joint Distribution}
\label{sec:2}
 In this section we extend the skip-gram with negative sampling (SGNS) word embedding algorithm \cite{Mikolov_nips}  to a general setup of embedding
   a discrete joint distribution. We also extend the algorithm analysis of Levy and Goldberg \cite{Levy_nips} and provide an explicit expression for   the quality of the PMI matrix approximation obtained by the embedding algorithm. We later show in Section \ref{sec:3} how this generic PMI approximation concept is applied to the task of language modeling and converted to the desired conditional probability estimates.

  Let $X$ and $Y$ be two random variables defined on alphabets $\mathcal{A}_X$ and $\mathcal{A}_Y$, respectively, with a  joint distribution $p(x,y)$.
We want to find  embeddings $\vec{x},\vec{y}\in \mathbb{R}^d$ for every $x\in \mathcal{A}_X$  and $y\in \mathcal{A}_Y$, that best reflect the joint distribution of $X$ and $Y$  in the sense defined below.
We can represent a given $d$-dimensional embedding by a $|\mathcal{A}_X| \times |\mathcal{A}_Y| $ matrix $m$
such that $m(x,y) =  \vec{x} \cdot \vec{y}$. The rank of the embedding matrix $m$ is (at most) $d$.

We define the score of a given embedding $m$ of $\mathcal{A}_X$ and $\mathcal{A}_Y$ to be:
\begin{equation} S(m)  =  \sum_{x,y} f_{x,y}(m(x,y)) =  \sum_{x,y} f_{x,y}( \vec{x} \cdot \vec{y})  \label{jscriterion}
\end{equation}
such that
\begin{equation}
 f_{x,y}(z) = \frac{1}{k\!+\!1} (p(x,y)\log \sigma(z) + k p(x)p(y) \log \sigma(-z)), \hspace{0.51cm} x\in \mathcal{A}_X, \, y\in \mathcal{A}_Y, \, z\in  \mathbb{R}
 \label{jscriterion2}
\end{equation}
where $\sigma()$ is the sigmoid function and $k$ is a positive integer.
The objective function $S(m)$ can be viewed as a log-likelihood function of a binary logistic regression classifier that treats a sample from a joint distribution $p(x,y)$ as a positive instance, and independent samples from the two marginal distributions as a negative instance, while $k$ is the proportion between negative and positive instances.
Given a fixed embedding  dimensionality $d$,  the optimal embedding $m$ is the one that maximizes the objective function $S(m)$.

The (shifted version of  the) pointwise mutual information (PMI) function is defined as:
\begin{equation}\mpmi(x,y)=\log \frac{p(x,y)}{p(x)p(y)}-\log k, \hspace{1cm} x\in \mathcal{A}_X, y\in \mathcal{A}_Y.
\end{equation}
We denote hereafter the matrix whose cell-entries are the pmi values as the PMI matrix.
It can be easily verified that  $f_{x,y}(z)$ is a concave function of $z$ and the global maximum is obtained at $z=\mpmi(x,y)$.
Hence, for each pair $(x,y) \in |\mathcal{A}_X| \times |\mathcal{A}_Y| $ we obtain that $f_{x,y}(\vec{x}\cdot\vec{y}) \le f_{x,y}(\mpmi(x,y))$.
Summing over all the pairs we obtain that the global optimum of the embedding score function is obtained at the PMI matrix:
\begin{equation}
S(m)  = \sum_{x,y}  f_{x,y} (m(x,y)) \le \sum_{x,y}  f_{x,y}(\mpmi(x,y)) = S(\mpmi).
\end{equation}
The optimal $d$-rank embedding matrix $m$ is the best $d$-rank matrix approximation of the PMI matrix.

We next derive an explicit description of the approximation criterion that quantifies the gap between $S(m)$ and $S(\mpmi)$.
For each matrix $m$ we define a distribution on $ \!\mathcal{A}_X \!\times \! \mathcal{A}_Y \times \{0,1\} \!  $:
$$ p_m(x,y,1) = q(x,y) \sigma (m(x,y)), \hspace{1cm}  p_m(x,y,1) = q(x,y) (1-\sigma (m(x,y)))$$
such that $q(x,y)$ is the following mixture distribution:
\begin{equation}
q(x,y) = \frac{1}{k\!+\!1}(p(x,y) + kp(x)p(y)), \hspace{1cm} x\in \mathcal{A}_X, \hspace{0.2cm} y\in \mathcal{A}_Y.
\end{equation}
Applying Bayes' rule we obtain:
\begin{equation}
p_{\mpmi}(1|x,y)= \sigma (\mpmi(x,y)) = \frac{ p(x,y)}{p(x,y)+kp(x)p(y)}
\label{pqxy}
\end{equation}
\noindent {\bf Theorem 1:} Every real valued $|\mathcal{A}_X| \times |\mathcal{A}_Y| $ matrix $m$ satisfies:
\begin{equation}
S(\mpmi) - S(m) =  \KL ( p_{\mpmi}(z|x,y) || p_m(z|x,y)) =
 \sum_{x,y,z} p_{\mpmi}(x,y,z) \log \frac{p_{\mpmi}(z|x,y)}{p_{m}(z|x,y)}.
\label{klexpression}
\end{equation}

\noindent {\bf Proof}:  It can be easily verified from Eq. (\ref{pqxy}) that:
  \begin{equation}
  \frac{p(x,y)}{k\!+\!1}= \frac{q(x,y) p(x,y)}{p(x,y)+kp(x)p(y)} =  q(x,y) \sigma (\mpmi(x,y)) = q(x,y) p_{\mpmi}(1|x,y)
 \label{defa1}
\end{equation}
 and in a similar way
  \begin{equation}
  \frac{kp(x)p(y)}{k\!+\!1}= q(x,y) p_{\mpmi}(0|x,y).
   \label{defa2}
\end{equation}
 The definition of the score function $S(m)$ (\ref{jscriterion}) implies that:
\begin{equation}
S(\mpmi) - S(m) = \frac{1}{k\!+\!1} \sum_{x,y} ( p(x,y)\log \frac{\sigma({\mpmi}(x,y))}{\sigma(m(x,y))} + k p(x)p(y) \log
\frac{\sigma(-{\mpmi}(x,y))}{\sigma(-m(x,y))} )
\label{defdiff}
\end{equation}
Substituting Eq. (\ref{defa1}) and (\ref{defa2})  in Eq. (\ref{defdiff}) yields:
\begin{equation}
  =\sum_{x,y} q(x,y) \sum_{z=0,1} p_{\mpmi}(z|x,y) \log \frac{p_{\mpmi}(z|x,y)}{p_{m}(z|x,y)}
 \end{equation}
 and finally using the definition of conditional KL divergence \cite{cover2012elements} we obtain:
 \begin{equation}
 =  \KL ( p_{\mpmi}(z|x,y) || p_m(z|x,y)). \Box
  \label{appcerterion}
 \end{equation}
 The definition of the joint distribution $p_m(x,y,z)$ implies that  marginal $(x,y)$ distribution is the same for all the matrices $m$: $$p_m(x,y)=p_{\mpmi}(x,y) = q(x,y).$$
 Hence, we can equivalently state Theorem 1 using un-conditional KL divergence:
 $$
 S(\mpmi) - S(m) =  \KL ( p_{\mpmi}(x,y,z) || p_m(x,y,z)).
 $$

Theorem 1 implies that  the optimal d-dimensional embedding (which maximizes the embedding score $S(m))$ minimizes the KL divergence between the distributions defined by the embedding matrix $m$ and the PMI matrix.  Note that since KL is always non-negative we obtain as a by-product another proof that the embedding score $S(m)$ (\ref{jscriterion})
 is maximized by the PMI matrix.

   The embedding problem we address here can be viewed as a matrix factorization of the PMI matrix
 that best reflects the joint distribution $p(x,y)$.
  Previous works have suggested other criteria for matrix factorization such as least-squares \cite{Eckart}
 and KL-divergence between the original matrix and the low-rank matrix approximation \cite{Lee}.
In our setup we look for the best low-rank  approximation of the  PMI matrix, based on the KL-divergence criterion stated in Eq. (\ref{klexpression}),
that best reflects the joint distribution of $X$ and $Y$.

Levy and Goldberg \cite{Levy_nips} showed  that SGNS's objective achieves its maximal value when for each word-pair $x,y$ the
inner product of the embedding vectors $\vec{x} \cdot \vec{y}=\mpmi(x,y)$. We derived here the same result for the general setting of embedding any joint distribution. The result in \cite{Levy_nips}, However, tells us nothing about the lower dimensional case where the embedding algorithm is actually interesting since at  that case the PMI matrix factorisation is forced to compress the joint distribution and thereby learn a meaningful embedding.  In contrast, we provided above an explicit KL divergence expression (\ref{klexpression}) for the low-dimensional matrix-factorization criterion that is optimized by the SGNS algorithm.

Assume that we do not know the exact joint distribution. We only have, instead,  a  set of samples $\{(x_t,y_t)\}$  from
  the joint distribution $p(x,y)$.  We can use a Montecarlo approximation of the expectation computed by the score (\ref{jscriterion}). To compute the first term of (\ref{jscriterion2}) we only go over the pairs $(x_t,y_t)$ that appear in the training set.
If the alphabet sizes are large, it is not feasible to compute the second term  of (\ref{jscriterion2}). Instead, we can approximate the expectation by sampling of `negative' examples.
The negative examples are created  for each
pair $(x_t,y_t)$ by drawing $k$ random samples from  the empirical marginal distribution $\hat{p}(y)$.
 The objective function (\ref{jscriterion}) is thus approximated as:
\begin{equation} S_{sgns}   =   \sum_{t}   (\log \sigma( \vec{x}_t \cdot \vec{y}_t )  +
   \sum_{i=1}^k   \log \sigma(- \vec{x}_t \cdot \vec{y}_i))
 \label{sgnsscore}
\end{equation}
such that $y_{i}$ are i.i.d. samples from the empirical marginal distribution $\hat{p}(y)$.

The SGNS word embedding algorithm  \cite{Mikolov_nips} aims to represent each word $x$ and each context word  $y$  as $d$-dimensional vectors $\vec{x}$ and $\vec{y}$ such that words that are ``similar"
to each other will have similar vector representations. Applying the objective function (\ref{sgnsscore})
to the word co-occurrence statistics,  we obtain the NEG objective function maximized by the SGNS algorithm.

In all the derivations above we were focused on approximating the PMI matrix. We can apply a similar analysis to obtain a  $d$-rank  matrix approximation of the log conditional distribution $\log p(y|x)$. Define
\begin{equation} S_{cond}  = \sum_{x,y} f_{x,y}( \vec{x} \cdot \vec{y}- \log (kp(x))).
\label{condopt}
\end{equation}
Then $S_{cond}$ is optimized when $\vec{x} \cdot \vec{y} -k \log p(x) =\mpmi(x,y)$, i.e. when $\vec{x} \cdot \vec{y}  = \log p(y|x)$.
Hence, the $d$-rank matrix that optimizes the objective function $S_{cond}$ is the  best $d$-rank matrix approximation of the log conditional distribution  matrix. 

\section{Language modeling based on PMI Approximation}
\label{sec:3}

In this section we apply the embedding algorithm described above to the joint distribution of a word $w$ and its left-side context $c$.
We utilize the connection between the optimal embedding and the PMI matrix factorization to construct  an approximation of the conditional distribution $p(w|c)$. As a result we obtain an efficient algorithm for learning a language model.

The joint distribution of a word sequence satisfies the chain rule:
\begin{equation}\log p(w_1,...,w_n)= \sum_{i=1}^n \log p(w_i|c_i)
\label{chain_rule}
\end{equation}
such that $c_i=(w_1,...,w_{i-1})$ is the left-side context of $w_i$.
We use a simple lookup table for the word representation and  an  LSTM recurrent neural network
to obtain a left-side context representation.
We train the word and left-side context embeddings to maximize the objective (\ref{sgnsscore}):
\begin{equation} S = \sum_{w,c} ( \log \sigma (\vec{w} \cdot \vec{c}) + \sum_{i=1}^k  \log \sigma (-\vec{u}_i \cdot \vec{c}))
\label{sgnscost}
\end{equation}
such that $w$ and $c$ go over all the words and contexts in a given corpus and  $u_1,...,u_k$ are words independently  sampled from the unigram distribution.
We showed above that by optimizing this objective we obtain the  best low-dimensional approximation of the PMI matrix associated with the joint distribution of the word and its context.
Hence, if the embedding dimensionality  is large enough we get a good approximation:
\begin{equation}  \vec{w} \cdot \vec{c} \approx \mpmi(w,c) = \log \frac{p(w|c)}{p(w)} -\log(k)
\label{sgns_approx}
\end{equation}
which yields the following estimation of the conditional distribution $p(w|c)$:
 \begin{equation}
\hat{p}(w|c) \propto  \exp(  \vec{w} \cdot \vec{c} )  p(w).
\label{exp_approx}
\end{equation}
Thus, we finally obtain a parametric language modeling, while avoiding the softmax operation over the entire vocabulary.
At test time we need to multiply the output of the parametric model by a unigram word distribution and apply softmax over the vocabulary to obtain a normalized conditional distribution.

Following  \cite{Mikolov_nips} we can sample negative instances from a smooth unigram distribution $p^{\alpha}(w)$ such that $0 \le \alpha \le 1$.
 By tuning $\alpha$, one can interpolate smoothly between sampling popular words, as advocated by the unigram distribution, and sampling all words
equally. Each $\alpha$ entails a different embedding. 
The optimized embedding based on negative sampling from a smooth unigram distribution  satisfies:
\begin{equation}
 \vec{w} \cdot \vec{c} \approx \log \frac{p(w|c)}{p^{\alpha}(w)} - \log{k}.
 \label{smoothapprox}
 \end{equation}

Note that once we set a value for $\alpha$ during training we also need to use the same value of $\alpha$  at test time to compute the conditional probability
$\hat{p}(w|c)$.

In our approach we use the same negative sampling approach used by the SGNS word embedding algorithm. 
To correctly use the learned embeddings for estimating conditional probabilities at test time we
multiply the parametric model by the unigram probability of the predicted word. We denote our approach of 
using negative sampling for language modeling as NEGLM.

\section{Connection to Noise Contrastive Estimation}
\label{sec:4}
The language modeling algorithms that are most related to our approach are  those that are based on the Noise Contrastive Estimation (NCE) principle \cite{Gutmann2012}. There are several applications of NCE to language modeling that mainly differ from each other by the neural network architecture used to produce a parametric representation for the left-side context of the predicted word \cite{Zoph2016} \cite{Vaswani2013} \cite{Chen2015}.
In this section we first briefly review the NCE algorithm in the context of language modeling and then explain the difference between the standard implementation of NCE and our approach, which can be viewed as a simplified variant of NCE.
NCE transforms the parameter learning problem into a binary classifier training problem.
Let $p(w|c)$ be the probability of a word $w$ given a context $c$ and let $p_n(w)$ be a
`noise'  word distribution (e.g. a unigram distribution). The NCE approach assumes that  the word $w$ is sampled from a mixture distribution $\frac{1}{k\!+\!1}(p(w|c)+kp_n(w))$ such that the noise samples are $k$ times more frequent than samples from the `true' distribution $p(w|c)$.
Let $y$ be a binary random variable such that $y=0$ and $y=1$ corresponds to a noise sample and a `true' sample, respectively, i.e.
$p(w|c,y=0)=p_n(w)$ and $p(w|c,y=1)=p(w|c)$.
Assume the distribution of word $w$ given a context $c$ has the following parametric form:
\begin{equation}
p_{\theta}(w|c) = 
\frac{1}{Z_c} \exp(\vec{w} \cdot \vec {c}+b_w)
\label{loglinearm}
\end{equation}
such that $\vec{w}$ and $\vec{c}$ are vector representations of  the word $w$  and its context $c$.
Applying Bayes rule it can be  easily  verified that:
\begin{equation}
 p_{\mbox{nce}}(y=1|w,c) =   \sigma ( \vec{w} \cdot \vec{c} +b_w - \log Z_c  - \log (k  p_n(w)))
 \label{ncescore}
\end{equation}
where $\sigma()$ is the sigmoid function.
NCE uses Eq. (\ref{ncescore}) as an objective to train a binary classifier that decides which distribution was used to sample $w$. For each pair $(w,c)$ from the training corpus we sample
$k$ words $u_1,...,u_k$ from the noise distribution $p_n(w)$ to create $k$ negative examples $(u_1,c),...,(u_k,c)$.
The objective function optimized by the NCE is:
\begin{equation}
S_{\mbox{nce}} = \sum_{w,c} ( \log p(y=1|w,c) + \sum_{i=1}^k  \log p(y=0|u_k,c))
\end{equation}
In principle, to find the model parameters,  we need to estimate the normalization factor function $Z_c$ for each context $c$.  However, it was found empirically \cite{Mnih2012} that setting $Z_c=1$ didn't hurt the performance (see also theoretical analysis in \cite{Andreas_2015}).
Chen et al. \cite{Chen2015} reported that setting $\log(Z_c)=9$ gave them the best results.
In any case, once the  NCE classifier is trained, at test time we still need to apply softmax (\ref{loglinearm}) over the vocabulary to obtain a normalized conditional distribution.

Mikolov et al. \cite{Mikolov_nips} suggested the negative sampling (NEG) training procedure, which is a  simplified version of the objective function optimized by NCE. The main difference between NEG and NCE is that NCE objective function comprises both
samples and an explicit  description of the noise distribution, while NEG uses only
samples. The binary classification score used by NEG  (for word embedding) as well as by our proposed NEGLM is:
 \begin{equation}
 p_{\mbox{neg}}(y=1|w,c) =   \sigma ( \vec{w} \cdot \vec{c}).
 \label{negscore}
\end{equation}
Comparing (\ref{ncescore}) and (\ref{negscore}), the NEG objective function is simpler and is a more stable input to the sigmoid function,
since it is easier to guarantee that the input to the sigmoid is concentrated around zero with low variance.
It can be viewed as an automatic implicit batch normalization of the input to the non-linear activation function.
If we consider the NEG objective function (\ref{negscore}) simply as an approximation of the NCE objective function, while using the same NCE formulation at test time we obtain poor results  (in terms of perplexity). However, unlike NEG, we apply the correct procedure at test time, which is  multiplying the output of the parametric model by the unigram word distribution. The three possible LM training strategies are  summarized in Table 1.

We can also view our approach not just as an approximation, but also as a special case of NCE by setting $b_w= \log (kp(w))$ in the NCE objective function (\ref{ncescore}).
We can address this in a systematic way.
 In the NCE we use a parametric  model for the `true' distribution $p_{\theta}(w|c)=\frac{1}{Z_c} \exp(\vec{w} \cdot \vec{c}+b_w)$.
      Assume that we instead apply NCE on a  parametric model for the quotient of the true distribution and the noise distribution:
 \begin{equation}
    \frac{p_{\theta}(w|c)}{p_n(w)}=\frac{1}{Z_c}\exp(\vec{w} \cdot \vec{c}).
    \label{quetient}
  \end{equation}
  Applying the  NCE formulation defined above on  (\ref{quetient}) we obtain:
  \begin{equation}
 p(y=1|w,c) =  \sigma ( \vec{w} \cdot \vec{c} - \log Z_c)
 \label{pmiscore}
\end{equation}
Setting $Z_c=1$ for all context $c$ we finally obtain the objective function we optimize (\ref{sgnscost}).
\begin{table}
\center
\caption{Comparison of the objective functions that are used for training and the obtained conditional word distribution between NCE, NEG and  NEGLM algorithms.}
\begin{tabular}{|l|l|l|}
  \hline
   & train & test \\\hline
    NCE & $p(y=1|w,c)=\sigma(\vec{w} \cdot \vec{c} + b_w -  \log (kp(w))) $ & $p(w|c) \propto \exp (\vec{w} \cdot \vec{c} +b_w) $ \\
  NEG & $p(y=1|w,c)=\sigma(\vec{w} \cdot\vec{c})$ & $p(w|c) \propto \exp (\vec{w} \cdot \vec{c}) $ \\
  NEGLM & $p(y=1|w,c)=\sigma(\vec{w} \cdot\vec{c}) $ &  $p(w|c) \propto \exp (\vec{w} \cdot \vec{c} + \log(p(w))) $  \\
  \hline
\end{tabular}
\end{table}

As mentioned above, it was found empirically \cite{Mnih2012} that setting $Z_c=1$ in the NCE objective function didn't hurt the performance.
The connection we established between NCE and matrix approximation can provide a simple explanation for that empirical observation.
The NCE can be viewed as an algorithm that finds the best low-rank approximation of the $\log p(w|c)$ matrix (see Eq. \ref{condopt}).
Assume we obtain the best $d$-dimensional NCE approximation with word and context embeddings  $\vec{w},\vec{c}\in R^d$ and a normalization factor
$Z_c=\sum_w \exp(\vec{w} \cdot \vec{c} + b_w)$. We can form a $(d\!+\!1)$-rank embedding  $(\vec{w},1),(\vec{c}, -\log(Z_c))\in R^{d\!+\!1}$ with $Z_c=1$ .  The NCE binary classier in both cases is $p(y=1|w,c)=\sigma(\vec{w} \cdot \vec{c} + b_w - \log(Z_c) -\log (kp(w)))$ and they yield the same language model.
The optimal $(d\!+\!1)$-rank NCE embedding  with normalization factor $Z_c=1$ yields a  better approximation of  the $\log p(w|c)$ matrix
than the $(d\!+\!1)$-rank NCE embedding $(\vec{w},1),(\vec{c}, -\log(Z_c))\in R^{d\!+\!1}$. Hence, a $d\!+\!1$-rank model with $Z_c=1$ 
provides a better conditional distribution approximation than a $d$-rank model with a properly defined $Z_c$.  
 Therefore, setting the normalization factor to 1 does not hurt the performance.

\section{Experiments}
\label{sec:5}

\label{subsec:lm}
We evaluate the performance of the language modeling methods discussed in this paper, using the popular perplexity measure on two standard datasets.
We denote the NCE language model as \emph{NCE}.
We use the heuristics that worked well in \cite{Vaswani2013}\cite{Zoph2016}, initializing NCE's bias term from Equation \ref{ncescore} to $b_{w} = -\log|V|$, where $V$ is the word vocabulary, and using $Z_{c} = 1$. We denote our proposed language model as \emph{NEGLM}. We also evaluate a variant of our model, denoted \emph{NEGLM-B}, where we adopt the learned bias term used in NCE. In this model $p(y=1|w,c)=\sigma(\vec{w} \cdot\vec{c} + b_{w})$ and the bias term $b_{w}$ is initialized to zero. We note that the goal of the evaluation described in this section, is to compare between the discussed language model variants under the same terms. We do not compare our results to state-of-the-art as we do not have sufficient computational resources to learn models that are large enough to be that much competitive. Neural network training typically requires making several hyperparameter choices. We describe these choices below and stress that for fair comparison, we followed prior best practices, while never optimizing these hyperparameters in favor of NEGLM.
All of the models described hereafter were implemented using the Chainer toolkit \cite{tokuichainer}.\footnote{We will make our code publicly available.}

The first dataset that we used is the Penn Tree Bank (PTB). A version of this dataset that is commonly used to evaluate language models is available from Tomas Mikolov.\footnote{\url{http://www.fit.vutbr.cz/~imikolov/rnnlm/simple-examples.tgz}} It consists of 929K training words, 73K validation words and 82K test words with a 10K
word vocabulary. To build and train all compared models in this setting, we followed \cite{zaremba2014recurrent}, which achieved excellent results on this dataset. Specifically, we used a 2-layer 300-hidden-units\footnote{\cite{zaremba2014recurrent} use larger models with more units.} LSTM with a 50\% dropout ratio, to represent the left-side context of a predicted word. We represent end-of-sentence as a special <eos> token and predict this token like any other word. During training we perform truncated back-propagation-through-time, unrolling the LSTM for 20 steps at a time without ever resetting the LSTM state. We train our model for 39 epochs using Stochastic Gradient Descent (SGD) with learning rate of 1, which is decreased by a factor of 1.2 after every epoch starting after epoch 6. We clip the norms of the gradient to 5 and use mini-batch size of 20. We set the negative sampling parameter to $k=100$ following \cite{Zoph2016}, which showed highly competitive performance with NCE language models trained with this number of samples.

As the second dataset, we used the much larger WMT 1B-word benchmark, \footnote{\url{http://www.statmt.org/lm-benchmark/1-billion-word-language-modeling-benchmark-r13output.tar.gz}} introduced by \cite{chelba2013one}. This dataset comprises about 0.8B training words and has a held-out set partitioned into 50 subsets. The test set is the first subset in the held-out, comprising 159K words, including the <eos> tokens. We used the second subset as our validation set with 165K words. The original vocabulary size of this dataset is 0.8M words after converting all words that occur less than 3 times in the corpus to an <unk> token. However, we follow \cite{williams2015scaling}\cite{ji2016blackout} and trim the vocabulary further to the top 64K most frequent words in order to successfully fit a neural model to this data using reasonably modest compute resources. To build and train our models we use a similar method to the one used with PTB, with the following differences. We use a single-layer 512-hidden-unit LSTM to represent the left-hand context. We follow \cite{jozefowicz2016exploring}, which found a 10\% dropout rate to be sufficient for relatively small models fitted to this large training corpus. We train our model for just one epoch using the Adam optimizer \cite{kingma2014adam} with default parameters, which we found to converge more quickly and effectively than SGD. We use mini-batch size of 1000.

\begin {table}[h]
\begin{center}
\caption{Perplexity results on test sets.}
\label{tab:results}
\begin{tabular}{|l|c|c|c|}
\hline
	 & NEGLM &  NEGLM-B & NCE \\
\hline
	PTB & \textbf{98.35}  & 100.69 & 104.33 \\
\hline
	WMT & \textbf{65.84} & \textbf{65.62} &  69.28 \\
\hline
\end{tabular}
\end{center}
\end {table}

The perplexity results achieved by the compared models on the two test sets appear in Table \ref{tab:results}. As can be seen, the performance of all models is comparable with NEGLM models outperforming the NCE model by a small margin. Furthermore, we see that our NEGLM model performs as well or slightly better than the one enhanced with a learned bias component (NEGLM-B). This suggests that the learning of our NEGLM model is robust as is without such a component.

\comment{
 \begin{figure}[ht]
\centering
 \includegraphics [scale=0.30]{ptb.png}
 \includegraphics [scale=0.30]{wmt2.png}
 \caption{Learning curves for the  PTB dataset (left) and the WMT-1B dataset (right).}
\label{learning_curves}
\end{figure}

To understand the empirical properties of the proposed algorithm, we next demonstrate via learning curves that negative sampling is more stable than NCE. Figure \ref{learning_curves} shows learning curves on two popular datasets (Penn Treebank and the
billion Word benchmark. The figure shows learning curves for several algorithms.
NCE-Z is when we training NCE model by initializing the learned bias to zero, and NCE-H is the heuristic initialization that we used .
NEGLM is our model and NEGLM-B is our model with a learned bias component added to it.
}

\section{Conclusions}

In this work we first derived an information-theoretic foundation for the relation between negative sampling and PMI approximation. This analysis extends previous analysis of Levy and Goldberg \cite{Levy_nips} to the case of lower dimensional PMI matrix approximation.
    We have shown that the simplified negative sampling (NEG) objective function, popularized in the context of learning word representations, can be used to learn parametric language models (NEGLMs) as well, as long as the correct procedure is followed at test time.
    Thus we provided  a unified approach based on PMI approximation for both word embedding and language modeling.
    Analyzing the relation between our proposed NEGLMs and NCE language models, we have shown that while they are very closely related, NEGLMs have the advantage of a simpler objective function, while NCE LMs require some heuristic measures to achieve such stability. Empirical evaluations showed that in out settings, NEGLMs slightly outperform NCE LMs on two popular perplexity measure benchmarks. As an additional contribution, we provide a formal derivation to how any discrete joint distribution can be approximated by means of learning continuous embeddings with the negative sampling objective function. We use this to derive our NEGLM model, but independently this also provides an alternative interpretation for the NCE learning method. 

\bibliographystyle{plain}
\bibliography{paper}
\end{document}